\documentclass[lettersize,journal]{IEEEtai}
\usepackage{amsmath,amsfonts,amssymb}
\usepackage[ruled,vlined,linesnumbered]{algorithm2e}
\usepackage{array}
\usepackage[caption=false]{subfig}
\usepackage{textcomp}
\usepackage{algorithmic}
\usepackage{verbatim}
\usepackage{graphicx}
\def\BibTeX{{\rm B\kern-.05em{\sc i\kern-.025em b}\kern-.08em
    T\kern-.1667em\lower.7ex\hbox{E}\kern-.125emX}}
\usepackage{balance}
\usepackage{booktabs}
\usepackage{diagbox}
\usepackage{makecell}
\usepackage{arydshln}
\usepackage{threeparttable}
\usepackage{url}
\usepackage{multirow}
\usepackage{pifont}
\usepackage{setspace}
\usepackage[capitalize]{cleveref}
\Crefname{section}{Section}{Sections}
\Crefname{table}{Table}{Tables}
\Crefname{equation}{Equation}{Equations}


\begin{document}
\title{Improve Knowledge Distillation via Label Revision and Data Selection}
\author{Weichao Lan, Yiu-ming Cheung, \IEEEmembership{Fellow, IEEE}, Qing Xu, Buhua Liu, Zhikai Hu, Mengke Li, Zhenghua Chen
% \author{Anonymous author
\thanks{
% This work is supported by the Agency for Science, Technology and Research (A$^*$STAR) Singapore under its NRF AME Young Individual Research Grant (Grant No. A2084c1067)

Weichao Lan, Yiu-ming Cheung, Buhua Liu and Zhikai Hu are with the Department of Computer Science, Hong Kong Baptist University, Hong Kong SAR, China (e-mail:\{cswclan, ymc, csbhliu, cszkhu\}@comp.hkbu.edu.hk).

Zhenghua Chen and Qing Xu are with the Institute for Infocomm Research, Agency for Science, Technology and Research (A*STAR), 138632, Singapore.(E-mail: chen0832@e.ntu.edu.sg, Xu\_Qing@i2r.a-star.edu.sg)

Mengke Li is with Guangdong Laboratory of Artificial Intelligence and Digital Economy (SZ), Guangdong, China (E-mail: limengke@gml.ac.cn).}
\thanks{Yiu-ming Cheung is the Corresponding Author.}
}

% \markboth{Submitted to IEEE Transactions on Artificial Intelligence}{W.C. Lan \MakeLowercase{\textit{et al.}}: Improve Knowledge Distillation via Label Revision and Data Selection}

\maketitle
\begin{abstract}
Knowledge distillation (KD) has become a widely used technique in the field of model compression, which aims to transfer knowledge from a large teacher model to a lightweight student model for efficient network development. In addition to the supervision of ground truth, the vanilla KD method regards the predictions of the teacher as soft labels to supervise the training of the student model. Based on vanilla KD, various approaches have been developed to further improve the performance of the student model.
% , such as extracting more informative knowledge using powerful feature representations or introducing a more efficient training strategy. 
However, few of these previous methods have considered the reliability of the supervision from teacher models. 
% It is worth noting that the teacher model may make incorrect predictions, despite performing well on specific tasks. 
Supervision from erroneous predictions may mislead the training of the student model. This paper therefore proposes to tackle this problem from two aspects: Label Revision to rectify the incorrect supervision and Data Selection to select appropriate samples for distillation to reduce the impact of erroneous supervision. In the former, we propose to rectify the teacher's inaccurate predictions using the ground truth. 
% Through such integration, wrong predictions can be revised while maintaining the relative information among different classes. 
In the latter, we introduce a data selection technique to choose suitable training samples to be supervised by the teacher, thereby reducing the impact of incorrect predictions to some extent. Experiment results demonstrate the effectiveness of our proposed method, and show that our method can be combined with other distillation approaches, improving their performance.
\end{abstract} 

\begin{IEEEImpStatement}
Knowledge distillation has significance and potential implications in the field of machine learning and model compression, making it a powerful technique for model optimization and deployment in real-world scenarios. It enables the transfer of knowledge from a large, highly accurate teacher model to a smaller, computationally efficient student model. This process not only reduces the model size but also improves its generalization capabilities.
However, in knowledge distillation, despite performing well on specific tasks, the teacher model may make incorrect predictions which can potentially mislead the training of the student model. In this paper, the proposed method aims to address the issue of erroneous predictions from the teacher model. By incorporating two key aspects, label revision and data selection, these approaches seeks to minimize the impact of incorrect supervision, with the potential to enhance the accuracy and reliability of knowledge distillation.
\end{IEEEImpStatement}

% \IEEEtitleabstractindextext{
\begin{IEEEkeywords}
Knowledge Distillation, Lightweight Model, Image Classification
\end{IEEEkeywords}
% }

\section{Introduction}
\label{sec:intro}
In recent years, lightweight models have attracted more and more attention for deploying deep neural networks (DNNs) on resource-constrained devices due to their property of being less parameterized \cite{han2015deep, howard2017mobilenets, zawish2024comp}. Among the various approaches used to build and train lightweight models, knowledge distillation (KD) has been proven to be a highly effective method for achieving model compression, and promoting the performance of lightweight models in various applications \cite{lin2022, xu2023rein, kao2023specific, wang2024gene}. KD works by transferring knowledge from a high-capacity network (the teacher) to a smaller one (the student) \cite{hinton2015distilling, wang2021knowledge, chen2022knowledge}. Typically, the teacher is a large neural network or network set with a large number of parameters, whereas the student network is compact and lightweight. Given a powerful teacher network, it is used to supervise the training of the student by utilizing the information (referred to as ``knowledge") such as final predictions \cite{hinton2015distilling, mirzadeh2020improved}, intermediate feature maps \cite{romero2014fitnets, jin2019knowledge}, or the relationships between different layers or samples \cite{lee2018self, passalis2020heterogeneous}. Under the teacher's supervision, the accuracy of the student network will be significantly improved, with much less storage and computation cost.

The seminal work of \cite{hinton2015distilling} trains the student using the logits of the teacher, which provides additional knowledge of inter-class probabilities and similarities. Specifically, the student is trained to mimic the predictions of teacher by minimizing the Kullback-Leibler (KL) divergence. In addition to the classical supervision of ground-truth labels, an extra logit loss is introduced as a powerful regularization, where the teacher's predictions are regarded as the soft labels to supervise the training of student. One drawback of this vanilla KD method is that the knowledge of the teacher is only represented by the final layer, ignoring the enriched intermediate-level information that has been proven to be crucial for learning representations \cite{romero2014fitnets, gou2021knowledge}. Recent works are built upon the theoretical basis of vanilla KD, aiming at further capturing the wealth of knowledge contained in intermediate feature maps by exploiting auxiliary loss functions to overcome this limitation \cite{tung2019similarity, ahn2019variational, chen2021cross, tian2019contrastive}. For instance, Yang et al. \cite{yang2021knowledge} introduced additional feature matching and regression losses to optimize the penultimate layer feature of the student. SemCKD \cite{chen2021cross} uses an attention mechanism to minimize the calibration loss in cross-layer knowledge distillation. There are also some methods that focus on extending different training strategy to improve transfer efficiency \cite{lee2018self, xu2020knowledge}. 
% In addition to studies on knowledge tpyes, some works also explore distillation spots such as SAKD \cite{song2022spot}, which decides the distillation spots adaptively, and  
Through these approaches, the performance of the student models has been constantly enhanced. 

\begin{figure*}[!t]
\centering
\includegraphics[width=1.7\columnwidth]{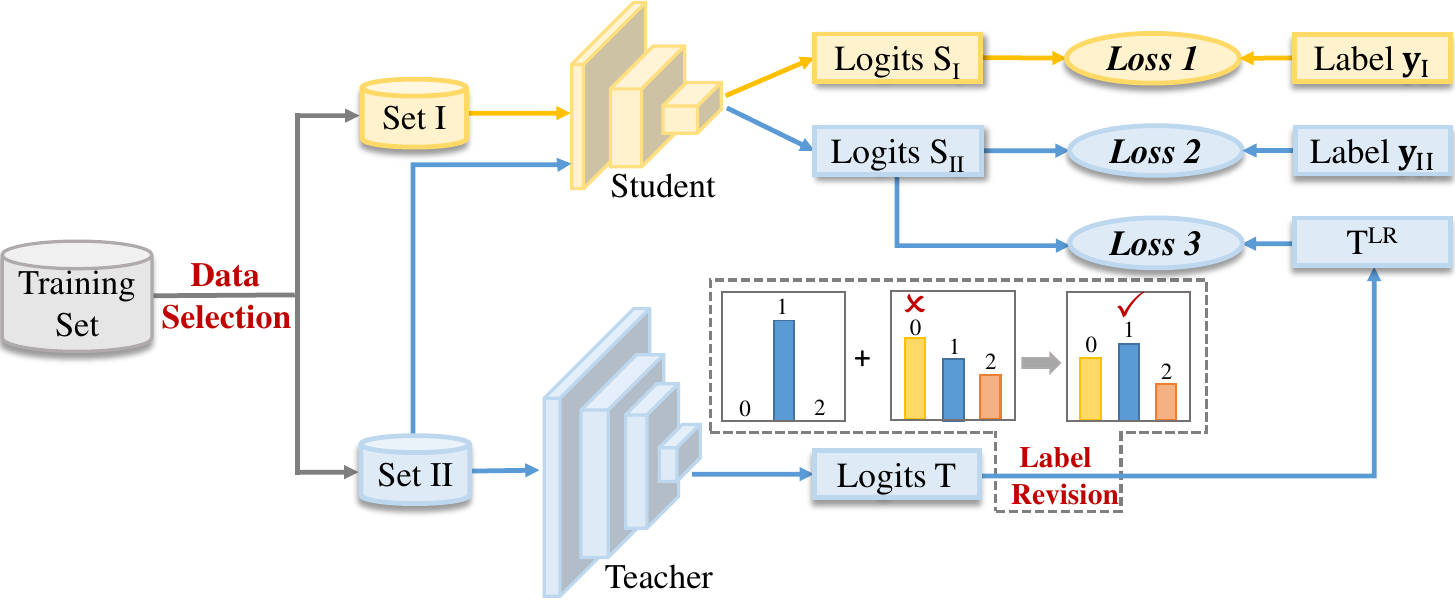}
\caption{An overview of training procedure with our proposed Label Revision (LR) and Data Selection (DS). For the entire training set, it is first split into two parts through DS, i.e., Set I and Set II. For Set I, it is input to the student model, then \textit{Loss 1} is calculated as the cross-entropy loss between the obtained Logits $\text{S}_\text{I}$ and Label $\text{y}_\text{I}$. With respect to Set II, it is input to the student and teacher at the same time, where the teacher's predictions Logits T is revised via LR before computing the distillation loss \textit{Loss 3}. Similar to \textit{Loss 1}, \textit{Loss 2} is the cross-entropy loss between Logits $\text{S}_\text{II}$ and Label $\text{y}_\text{II}$. Finally, the whole loss is a weighted summation of the three losses. More details about the calculation of losses are provided in Section \ref{sec:method}.}
\label{framework}
\end{figure*}

% Although the performance of student has been constantly enhanced through these methods, most of the algorithms are developed by introducing extra loss based on the vanilla KD losses, including the cross-entropy loss between the student's predictions and ground-truth labels, and the logits loss between the logits of student and teacher. However, these previous works have not given much consideration to the reliability of the teacher's logits.

% where the loss functions usually contains three parts. The first part is the ordinary loss such as cross-entropy between student predictions and ground-truth labels, and the second part comes from the difference (i.e., KL or MSE loss) between student and teacher predictions, then an auxiliary loss will be introduced according to different methods. Finally, the whole loss function can be expressed as the weighted summation of these three losses. 
% In this paper, we revisit the first two losses that has been applied in most of the KD methods. 

However, most of the algorithms are developed based on the vanilla KD framework, under the supervision of teacher models. Although the teacher usually has been trained well on specific tasks, it still contains incorrect knowledge such as wrong predictions. The previous works have not given much consideration to the reliability of the teacher models.
Since the ground truth labels (hard labels) and teacher predictions (soft labels) are both utilized to supervise the student in vanilla KD, two problems will arise naturally during the learning process. The first problem is that the teacher will also assign probabilities to incorrect classes. Although this kind of ``dark knowledge'' typically contains information on relative probabilities that has been shown to be beneficial for generalization, it is not fully trustworthy and some incorrect knowledge such as wrong predictions will also be transferred to student, misleading the direction of learning \cite{xu2020knowledge}. Additionally, wrong predictions will contradict ground truth that may cause confusion. Analogous to real-life learning in classes, the students will not be able to make correct judgments if they receive inconsistent information from different teachers for the same task. Therefore, to alleviate the negative impact of such incorrect supervision from the teacher and contradictions with ground truth, we first propose Label Revision (LR) to rectify the wrong predictions from the teacher via ground truth. Specifically, the ground-truth label is first reformulated as one-hot label, and then the soft labels of teacher are combined with the one-hot label based on meticulously designed rules. In this way, the wrong probabilities in teacher predictions can be revised, while the relative information among different classes can also be maintained. 
 
The other concern is whether the student needs the supervision from teacher on the entire dataset. Intuitively, the more guidance provided by the teacher, the greater the probability of containing wrong predictions. Therefore, we further introduce a Data Selection (DS) technique to select appropriate data for distillation, thereby reducing the impact of incorrect supervision to some extent. During the training of the student model with a whole training set, only a portion of the training samples are selected to be supervised by the teacher with logits loss, while the remaining samples are directly supervised by ground-truth using a single cross-entropy loss. The entire procedure of the proposed method, including LR and DS, is illustrated in Figure \ref{framework}. First, the entire training set is split into two parts (i.e., Set I and Set II) based on a certain criterion. Next, Set I is input to both the teacher and student models to obtain the logits loss, where the teacher's predictions are revised by LR before calculating the loss. For Set II, it is only input to the student and the loss is computed as the cross-entropy loss between student's logits and ground truth. 
We summarize our main contributions as follows:
\begin{itemize}
    \item [1)] To obtain more reliable supervision from the teacher model, we propose to rectify the incorrect predictions contained in teacher's soft labels using the ground truth. Without destroying hidden knowledge, the revised soft labels still maintain the relative information among different classes.
    \item [2)] We also introduce a data selection technique to select appropriate training samples for distillation from the teacher to the student, which further decreases the impact of wrong supervision.
    \item [3)] Extensive experiments with different datasets and network architectures are conducted to illustrate the effectiveness of our proposed method. It is also demonstrated that our method can be applied to other distillation approaches and bring improvements on their performance.
\end{itemize}

% \begin{figure}[t]
% \centering
% \includegraphics[width=1\columnwidth]{Figure/label_figure.pdf}
% \caption{Network construction.}
% \label{frame}
% \end{figure}

% a plug-and-play filter
% Label Revision (LR) and Data Selection(DS)
% from two aspect to decrease the effect of wrong prediction:
% 1. revise wrong soft labels
% 2. some samples do not need the supervision of teacher--data selection
\section{Related Work}

\subsection{Knowledge Distillation}
The concept of knowledge distillation was first introduced in the pioneering work of Hinton \emph{et al.} \cite{hinton2015distilling}. This technique leverages the knowledge contained in a larger teacher model to train a smaller student model. Initially, the method regards the output of the softmax layer of the teacher network, which is referred to as logits, as soft labels to supervise the training of the student. Building upon this idea, subsequent works have introduced various knowledge representations to achieve more accurate and efficient learning. These methods can be roughly divided into three categories according to the type of knowledge: logits-based distillation, feature-based distillation, and relation-based distillation \cite{gou2021knowledge}.

\textbf{Logits-based Distillation}
As used in \cite{hinton2015distilling}, logits refer to the final predictions of models. The vanilla method converts the logits into soft probability through a softmax function, where a temperature parameter is introduced for scaling small probabilities that contain valuable information and play an important role in supervision. To further improve accuracy, numerous methods have explored the potential of soft labels and made better use of logits. Zhou \emph{et al.} \cite{zhou2021rethinking} analyzed the impact of soft labels on the bias-variance trade-off during training and proposed to dynamically distribute weights to different samples for balance. Mirzadeh \emph{et al.} observed that the performance of student network will greatly degrade when the gap between the student and teacher is large. Thus, they introduced a small assistant network to distill the knowledge step by step. Kim et al. \cite{kim2021comparing} explored mean squared error (MSE) to replace the original KL divergence loss in vanilla KD and achieved superior distillation performance. In SimKD \cite{chen2022knowledge}, the classifier of teacher model is reused to make predictions for student model, where the output of the student is scaled by additional layers to match the dimensions of the teacher's output. However, the deployment cost will also increase due to the additional layers and teacher's classifier. Instead of introducing extra components, Zhao \emph{et al.} \cite{zhao2022decoupled} directly reformulated the loss of vanilla KD, which identified the target class and non-target classes in classification tasks to balance the contributions of training samples more effectively.

\textbf{Feature-based Distillation}
Since the performance gap between the student and teacher models is still large after logits-based distillation, new representations of knowledge have been explored by leveraging the information of features. The intermediate layers play an important role, especially when the neural network is deep, and the information contained in these layers can also be utilized as knowledge to train the student. FitNets \cite{romero2014fitnets} first forces the student to mimic the corresponding intermediate features of teacher. Rather than utilizing the feature information, \cite{zagoruyko2016paying} further fitted the attention maps of the student and teacher. Besides, \cite{huang2017like} extended the attention map by neuron selectivity transfer. To make it easier to transfer the knowledge from the teacher to student, Kim \emph{et al.} \cite{kim2018paraphrasing} introduced several factors to represent the features in a more understandable format. Jin \emph{et al.} \cite{jin2019knowledge} proposed hint learning with a constraint on the route, which supervised the student by the outputs of hint layers in the teacher.
% Heo \emph{et al.} \cite{heo2019comprehensive} took feature position and distance function into account, proposing a new loss function to deal with the redundant information. 
VID \cite{ahn2019variational} transfers knowledge by maximizing variational information. Inspired by contrastive learning, CRD \cite{tiancontrastive} utilizes the representational knowledge in the teacher to capture the relationship among each dimension. 
CTKD \cite{zhao2020highlight} applies collaborative
teaching and trains the student using two teachers synchronously.
% KR \cite{chen2021distilling} uses multi-layer feature information of the teacher to supervise the learning of a single layer in the student at the same time. 
SRRL \cite{yang2021knowledge} focuses on training the student’s penultimate layer by using teacher's classifier. SemCKD \cite{chen2021cross} introduces an attention mechanism that can automatically assign the most semantically related layer in the teacher model for each student layer, and the work is further extended to different scenarios \cite{wang2022semckd}.
Kao \emph{et al.} \cite{kao2023specific} proposed SEL that enables the student to obtain various expertise knowledge from different networks.
% KDCT \cite{tan2022improving} improves the distillation performance with a customized teacher model.
% Yang \emph{et al.} \cite{yang2022knowledge} proposed to learn more meaningful feature knowledge through a self-supervision augmented distribution.
% Zhang \emph{et al.} \cite{zhang2021student} proposed evolutionary knowledge distillation, where the teacher and student are trained synchronously. 
% MGD \cite{yang2022masked} proposes to use part of the pixels to reconstruct the features of the teacher, instead of mimicking the features directly. 
% SKD \cite{yang2023skill} introduces meta-learning to capture and predict the behaviors in hidden layers of the teacher.

Compared with logits-based distillation, feature-based methods can capture richer information, while the computation cost will also increase due to heavy feature transformation.

\textbf{Relation-based Distillation}: Relation-based distillation methods focus on the relationship between different data samples or network layers. Lee \emph{et al.} \cite{lee2018self} proposed to use the correlation between feature graphs as knowledge and extract the key information via singular value decomposition. 
RKD \cite{park2019relational} transfers the structured relationship between the outputs of the teacher to student.
Passalis \emph{et al.} \cite{passalis2020heterogeneous} explored the hint information, which utilizes the mutual information flow from pairs of hint layers in the teacher to train the student. In addition to the relationship among different layers, data samples also contain rich knowledge. For example, Passalis \emph{et al.} \cite{passalis2020probabilistic} modeled the relationship of data samples as a probability distribution via the feature representation of the data, where the teacher and student can be matched by transferring the probability distribution. MASCKD \cite{gou2022multilevel} explores more powerful relation knowledge and introduce attention maps to build correlations between samples. 
% CAMD \cite{liu2023new} utilizes the relationship in the the features of teacher as supervision to optimize the student features.

\subsection{Data Selection}
When training a model, different training samples will have varying contributions to specific behaviors of model. Measuring the effect of different samples and selecting more appropriate ones for training can help improve the performance of model to some extent \cite{tonevaempirical, lin2022measuring}. In fact, data selection technique has been explored in many fields such as active learning \cite{paul2017non,liu2021influence}, adversarial learning \cite{sinha2019variational, wang2020dual}, transfer learning \cite{ruder2017learning, xiong2020source} and reinforcement learning \cite{yoon2020data}. To quantify the value of data samples for further selection, various algorithms have been explored. Koh \emph{et al.} \cite{koh2017understanding} estimated the effect of individual data through influence function \cite{cook1982residuals}, which perturbs each training sample and measures changes in the model's output. Data Shapley \cite{ghorbani2019data} regards the improvement of marginal performance as data values, where all the possible subsets of the whole training set are considered. TracIn \cite{pruthi2020estimating} tracks the gradient information during training and monitors the changes in model predictions as each training sample is accessed. 
% Ren \emph{et al.} \cite{ren2023single} introduced a Shapley-enhanced approach to weight the information. 
By quantifying the contributions of training data, more appropriate samples can be selected to further improve the performance or efficiency of model training. 
% Although data selection have been applied in various fields, ... 

\section{Proposed Method} \label{sec:method}

In this section, we first provide a brief introduction to the vanilla KD method including some essential notations. Then, we introduce the proposed method to revise soft labels of the teacher and describe the data selection technique applied before distillation in detail. For reference, the main notations are listed in Table \ref{notation}.

\begin{table}[t]
\small
\centering
\caption{Main notations.}
\centering
\renewcommand\arraystretch{1.2}
\setlength\tabcolsep{5pt}
\begin{tabular}{lp{7cm}}
% \hline
\bottomrule[0.7pt] \specialrule{0em}{1pt}{1pt}
$D$ & The entire training set, where $D = \{\mathbf{x}^{(i)},\mathbf{y}^{(i)}\}_{i=1}^N$. $\mathbf{x}$ and $\mathbf{y}$ are samples and labels, respectively.  \\
$D^t, D^s$ & Training subsets, where $D^t, D^s \subset D$.\\
$\Theta$ & Model parameter set.\\
$\theta^*, \hat{\theta}$   & Optimal model parameters, where $\theta^*, \hat{\theta} \in \Theta$.\\  
% \hline
$z^t, z^s$ & The logits of the \textbf{t}eacher and \textbf{s}tudent, respectively.\\
$z^s_r, z^s_w$ & The \textbf{r}ight and \textbf{w}rong part of \textbf{s}tudent's logits, respectively, which are split according to teacher's prediction.\\
$\epsilon^{(i)}$ & Perturbation added on sample $\mathbf{x}^{(i)}$.\\
$p^t$ & Softmax probabilities of the teacher.\\
$\sigma(\cdot)$ & The softmax function.\\
$\mathcal{L}(\mathbf{x},\theta)$ & Risk of model with parameter $\theta$ on sample $\mathbf{x}$.\\
$\mathcal{L}_r, \mathcal{L}_w$ & The loss of \textbf{r}ight and \textbf{w}rong part, respectively.\\
$\lambda_1, \lambda_2, \eta$ & Hyper-parameters.\\

\bottomrule[0.7pt]
\end{tabular}
\label{notation}
\end{table}

\subsection{Preliminaries}
\textbf{Vanilla Knowledge Distillation.} The fundamental concept behind vanilla KD is to train the student model to mimic the outputs of the teacher model, by minimizing the difference between their predictions for a given set of input data. This is accomplished by combining the original loss function (i.e., cross-entropy loss) with an additional distillation loss term. Suppose the logits of the teacher and student model are $z^t$ and $z^s$, respectively. Then, the total loss function of vanilla KD can be expressed as:
\begin{equation}\label{eq:vanilla_loss}
\mathcal{L} = \mathcal{L}_{CE}(\sigma(z^s), \textbf{y}) + \lambda_1 
\mathcal{L}_{KD} (z^s, z^t), 
\end{equation}
where $\mathcal{L}_{KD}$ donates the distillation loss and $\mathcal{L}_{CE}$ is the cross-entropy loss between model predictions and ground-truth label $\textbf{y}$ in classification. $\sigma(\cdot)$ refers to the softmax function.
Here, a hyperparameter $\lambda_1$ is introduced as a weight to balance the two losses. 

In general, the distillation loss $\mathcal{L}_{KD}$ is typically a soft version of the original loss function, which encourages the student to learn the same underlying information as the teacher model. As used in \cite{hinton2015distilling}, it is defined as the KL divergence between the logits of student and teacher. In the original softmax function, the output values are transformed into a probability distribution over the set of possible classes. To allow the student to capture more knowledge contained in teacher, the distillation loss involves a temperature parameter $\tau$ to the softmax function. Thus, the distillation loss $\mathcal{L}_{KD}$ can be reformulated as:
\begin{equation}\label{eq:kl_loss}
\mathcal{L}_{KD} = \tau^2\mathcal{L}_{KL} (\sigma(z^s/\tau), \sigma(z^t/\tau)).
\end{equation}
A higher temperature $\tau$ will result in a softer and more diffuse probability distribution. This softer distribution encourages student to learn from the decision-making process of teacher rather than simply mimicking its outputs. 

% The coefficient T2 is used to ensure that the gradient magnitude of the LKL part keeps roughly unchanged when the temperature is larger than one.

\textbf{Data Impact Estimation.}
There has been various algorithms to estimate the impact of different training samples that can help select data, such as influence-based methods and shapley-based methods. Among these methods, the influence function is one of the most popular tools for data selection due to the advantage of low complexity without retraining models \cite{hammoudeh2022training,li2021privacy}. The influence function is first developed in the field of statistics \cite{cook1982residuals,huber2011robust}, and has been applied to measure the influence of data samples \cite{koh2017understanding}. Without retraining the model, it provides an efficient way to estimate the change on model predictions or parameters if a sample is perturbed slightly. Let $D = \{\mathbf{x}^{(i)},\mathbf{y}^{(i)}\}_{i=1}^N$ donate a set of $N$ data pairs, and model parameter set is $\Theta$, where the optimal parameter after convergence is,
\begin{equation}
    \theta^* = \arg \underset{\theta \in \Theta} \min \frac{1}{N}\sum_{\mathbf{x} \in D}\mathcal{L}(\mathbf{x},\theta).
\end{equation}
When the i-th training sample $\mathbf{x}^{(i)}$ is perturbed by an infinitesimal step $\epsilon^{(i)}$, the new optimal parameters will change to 
\begin{equation}
    \hat{\theta} = \arg \underset{\theta \in \Theta} \min \frac{1}{N}\sum_{\mathbf{x} \in D}\mathcal{L}(\mathbf{x},\theta)+ \epsilon^{(i)}\mathcal{L}(\mathbf{x}^{(i)},\theta).
\end{equation}

Using influence function \cite{cook1982residuals}, the change on model parameters can be roughly estimated as:
\begin{equation}\label{eq:param_change}
\frac{d\hat{\theta}}{d\epsilon^{(i)}}\big|_{\epsilon^{(i)} =0} = -H_{\theta^*}^{-1}  \nabla_{\theta} \mathcal{L}(\mathbf{x}^{(i)}, \theta^*),
\end{equation}
where $H_{\theta^*} \triangleq \frac{1}{N}\sum_{\mathbf{x} \in D}\nabla_{\theta}^2 \mathcal{L}(\mathbf{x}, \theta^*)$ is the Hessian matrix and $\mathcal{L}(\mathbf{x}^{(j)}, \theta^*)$ is the loss at point $\mathbf{x}^{(j)}$. 

We can also approximate the change in model predictions at a test sample $\mathbf{x}^{(j)}$ based on the chain rule \cite{koh2017understanding}, that is, 
\begin{equation}\label{eq:pred_change}
\begin{split}
\frac{d\mathcal{L}(\mathbf{x}^{(j)}, \theta^*)}{d\epsilon^{(i)}}\big|_{\epsilon^{(i)} =0}
&=\frac{d\mathcal{L}(\mathbf{x}^{(j)}, \theta^*)}{d\hat{\theta}} \frac{d\hat{\theta}}{d\epsilon^{(i)}}\big|_{\epsilon^{(i)} =0} \\
&= -\nabla_{\theta} \mathcal{L}(\mathbf{x}^{(j)}, \theta^*) H_{\theta^*}^{-1}  \nabla_{\theta} \mathcal{L}(\mathbf{x}^{(i)}, \theta^*).
\end{split}
\end{equation}

Then, using Eq. (\ref{eq:param_change}) and (\ref{eq:pred_change}), the influence score of each sample on model parameters or predictions can be obtained. Further, we can select more appropriate data for training according to the scores.

\subsection{Revise Soft Labels of Teacher}

The loss function in Equation (\ref{eq:vanilla_loss}) shows that the student model is supervised by the logits of teacher $z^t$ and hard labels (i.e., $\textbf{y}$) simultaneously. However, even if the teacher model is pre-trained well, it can still make incorrect predictions, which may conflict with the guidance provided by the hard labels, leading to decline on accuracy of the student model. The vanilla KD method utilizes an extra cross-entropy loss to decrease the impact of wrong predictions, but the revision is insufficient to make reliable supervision and there remains noteworthy wrong information. To address this issue, we propose to improve the reliability of the teacher's supervision by revising its soft labels. 
Specifically, we focus on the wrong soft labels provided by the teacher and propose to revise them via hard labels, where the revised label is a linear combination of hard labels in one-hot form and teacher's soft labels. Assuming that the finally predicted probabilities of teacher after softmax is $p^t = \sigma(z^t) = [p^t_1, p^t_2,\dots,p^t_C]$, where $C$ is the number of classes. Then, the new revised soft label can be computed as: 
\begin{equation}\label{eq:revise}
p = \beta p^t +(1-\beta)\textbf{y}.
\end{equation}
 We donate $p^t_{max}$ as the corresponding probability of the predicted class, that is, the maximum probability, and $p^t_{tar}$ as the corresponding probability of the true target class. To rectify the probability so that the maximum probability is consistent with the hard label, the choice of weight parameter $\beta$ needs to meet the constraint, that is,
\begin{equation}\label{eq:beta}
\beta \times p^t_{tar} + (1-\beta) \times 1  > \beta \times p^t_{max} + 0
\Rightarrow
\beta < \frac{1}{p^t_{max}-p^t_{tar}+1}
\end{equation}
The constraint can be reformulated as,
 \begin{equation}\label{eq:eta}
\beta = \frac{\eta}{p^t_{max}-p^t_{tar}+1},
\end{equation}
where $\eta$ is a coefficient that is smaller than 1. Instead of directly swapping the predicted probability of the target class and incorrect class, as applied in \cite{wen2021preparing}, our proposed strategy can maintain the relative probabilities between similar classes from the perspective of feature representation, which has been proved to be beneficial to generalization of networks \cite{xu2020knowledge}. 
% To simplify the selection procedure, we introduce another parameter $\eta$ to calculate $\beta$ 

% 从特征角度分析，相当于交换了两类的特征, 他们考虑了抑制相似样本，但是忽略了xxx
% 我们考虑了利用相似样本信息，取得了一定效果
% An interesting and inspiring observation is that despite the teacher model assigns probabilities to incorrect classes, the relative probabilities of incorrect answers are exceptionally informative about generalization of the trained model.
% refer to: Preparing lessons: Improve knowledge distillation with better supervision

\begin{figure}[t]
\centering
\includegraphics[width=0.9\columnwidth]{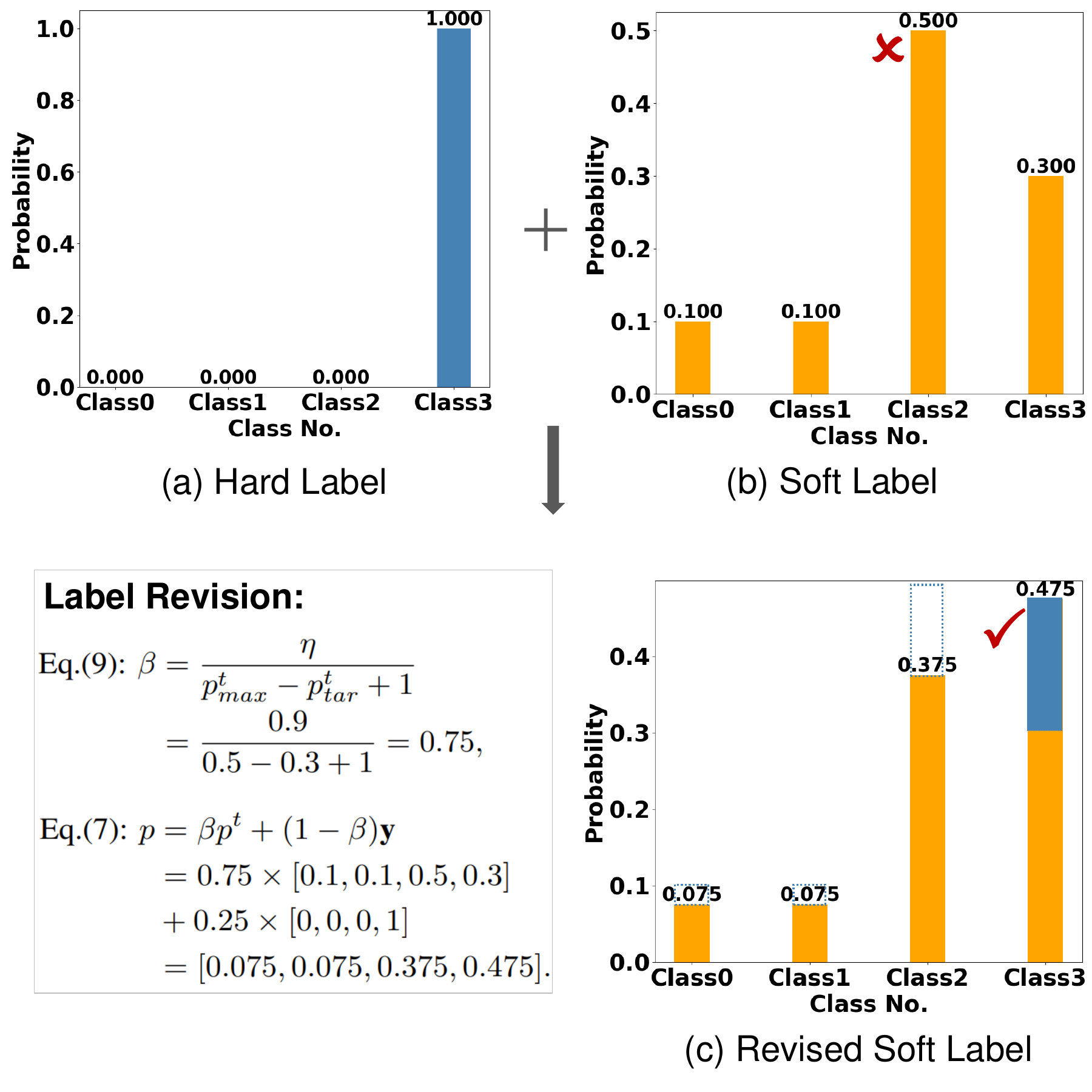}
\caption{A four-class example of Label Revision.}
\label{revise_label}
\end{figure}

Considering a simple case of a four-class classification, suppose that there is a sample belonging to Class 3 and the one-hot hard label is $\textbf{y}$ = [0, 0, 0, \textbf{1}]. However, the teacher model makes a wrong prediction as Class 2 with probabilities of $p^t$ = [0.1, 0.1, \textbf{0.5}, 0.3]. According to the proposed strategy of Eq. (\ref{eq:revise}), the soft label with be rectified as $p$ = [0.075, 0.075, 0.375, \textbf{0.475}] if $\eta$ is set as 0.9. As shown in Figure \ref{revise_label}, the target class obtains the maximum probability after revising. 

After obtaining the new soft labels, we can then compute the loss between student and teacher. To take full advantage of logits information, we replace the original KL loss with Mean Square Error (MSE) as suggested in \cite{romero2014fitnets,kim2021comparing}. Thus, the new loss of wrong predictions will be:
\begin{equation}\label{eq:wrong_part}
\mathcal{L}_w = \mathcal{L}_{MSE} (\sigma(z^s_w), p_w).
\end{equation}
The cross-entropy loss between ground truth and wrong predictions is not considered here, because $p_w$ has already contained the correct information from the true labels. The logits of the student are also normalized using the softmax function to match $p_w$. For the remaining right predictions that do not require revision, the loss $\mathcal{L}_r$ is calculated by Eq. (\ref{eq:vanilla_loss}). By integrating the wrong and right parts, we can then obtain the total new loss as:
\begin{equation}\label{eq:wrong_right}
\begin{split}
\mathcal{L} = & \mathcal{L}_r + \lambda_2\mathcal{L}_w \\
        = & \mathcal{L}_{CE} (\sigma(z^s_r), \textbf{y}_r)+\lambda_1\mathcal{L}_{MSE} (z^s_r, z^t_r) \\
        & + \lambda_2 \mathcal{L}_{MSE} (\sigma(z^s_w), p_w),
\end{split}
\end{equation}
where \{$\lambda_1, \lambda_2$\} are the coefficients to balance each term, which can be adjusted flexibly. The revised soft label of the wrong part $p_w$ is calculated by Eq. (\ref{eq:revise}).
% \lambda_3
% \begin{equation}\label{eq:wrong_part}
% \mathcal{L}_n = \underbrace{\mathcal{L}_{MSE} (z^s_r, z^t_r) +\mathcal{L}_{CE} (\sigma(z^s_r), \textbf{y})}_{\text{Right part} 
%  \mathcal{L}_r} + \underbrace{\mathcal{L}_{MSE} (z^s_w, \textbf{y}^n)}_{\text{Wrong part} 
%  \mathcal{L}_w}
% \end{equation}

\subsection{Select Appropriate Data to Distill}
Most previous KD methods concentrate on extracting more information from the teacher to improve the accuracy of the student, but few of them consider the impact of training data, which has a significant effect on the performance of supervised learning. To alleviate the wrong supervision from teacher, we have proposed to revise the soft labels in the previous section. From a different perspective of training data, it is also worth considering whether the supervision of the teacher is necessary to be applied to the entire training set. Therefore, we introduce a data selection technique to further decrease the risk of incorrect supervision.  

To train a student model, we select only a portion of the training samples to transfer the knowledge from the teacher, the remaining samples are directly supervised by ground-truth. Specifically, for a pre-trained teacher model, we first calculate the influence score of each sample in training set $D$ using Eq. (\ref{eq:param_change}) and sort the samples based on these scores to form a new set $D^n$. Then, we split $D^n$ into two subsets $D^s$ and $D^t$, where $D^t$ is supervised by the teacher with revised soft labels while $D^s$ is only supervised by ground-truth. The loss on subset $D^t$ is calculated using Eq. (\ref{eq:wrong_right}), while the loss on subset $D^s$ is the original cross-entropy loss between the student's prediction and ground-truth. Finally, the overall loss is the summation of these two losses. It is worth noting that the split of $D^n$ is flexible, for example, we can select the top 50\% (or 20\%, 80\% and so on) of the samples as $D^t$ and the remaining portion as $D^s$. In the experimental section, we conduct an ablation study to evaluate the performance of different split strategies.

To make a comprehensive view of our proposed method of LR and DS, we
summarize the whole process to train the student model in Algorithm 1.

\begin{algorithm}[t]
\caption{Knowledge Distillation with LR and DS.}
\begin{algorithmic}

\STATE \textbf{Input}: Training set $D=\{\mathbf{x},\mathbf{y}\}$; A pre-trained teacher model; A randomly initialized student model.
\STATE \textbf{Output}: A student model with better performance.
\end{algorithmic}
\begin{algorithmic}[1]
    \STATE \underline{\textbf{DS}}: For each example $\textbf{x}$ in $D$, compute the score based on Eq. (\ref{eq:param_change}) and obtain the new set $D^n$ after sorting.
    \STATE Split $D^n$ into $D^t$ and $D^s$.
	\FOR{iter = 1 to maxIter}
	    \STATE Get a combined minibatch of training data in both $D^t$ and $D^s$.
            \STATE \underline{\textbf{LR}}: For data pair in $D^t$, compute the logits of the teacher and revise it using Eq. (\ref{eq:revise}).
            \STATE Compute the loss of student on $D^t$ with Eq. (\ref{eq:wrong_right}).
            \STATE Get the cross-entropy loss on $D^s$ between ground truth and the student's logits.
            \STATE Update the parameters of student through backward propagation the gradients of total loss on $D^t$ and $D^s$.
	
    \ENDFOR
\end{algorithmic}\label{alg:LR_DS}
\end{algorithm}

\section{Experiments} \label{sec:exper}
To demonstrate the effectiveness of our proposed method, we conduct various experiments on image classification tasks. Firstly, we present the results of using or not using LR and DS, compared with vanilla KD. Next, we compare the performance our method with state-of-the-art distillation approaches. We also apply our method to other approaches to show that our method is compatible with them and can help improve their performance. Finally, ablation studies are also conducted to illustrate the sensitivity of our method to different settings of hyper-parameters. 

\subsection{Experiment Setup}
\subsubsection{Dataset and Network Architectures}

The experiments involve three popular datasets on classification tasks.

\textbf{CIFAR-100} \cite{krizhevsky2009learning}. The CIFAR-100 dataset consists of 100 classes with a total of 60K $32 \times 32$ colorful images, where each class has 50K training images and 10K test samples.

\textbf{ImageNet-2012} \cite{deng2009imagenet}. ImageNet-2012 is a large-scale dataset that contains around 1.2M training images and 50K validation samples from 1,000 different classes. The sizes of images in ImageNet are various so that they are often cropped as 224x224 for uniform. 

% \textbf{Tiny-ImageNet} \cite{hansen2015tiny}. As a subset of ImageNet, Tiny-ImageNet selects 200 classes and 600 images for each class, including 500 training images, 50 for validation and 50 testing images. The images in Tiny-ImageNet are downsapled to 64x64.

With respect to the network architectures, different combinations of teacher-student model are applied for evaluation. The selected architectures are all widely-used in classification, that are, VGG \cite{simonyan2014very}, ResNet \cite{he2016deep}, ShuffleNet \cite{ma2018shufflenet, zhang2018shufflenet} and MobileNet \cite{howard2017mobilenets, sandler2018mobilenetv2}.

\subsubsection{Compared Methods}
We compare our proposed method with both logit-based and feature-based methods. We also combine our method with other distillation approaches to illustrate that the proposed method is compatible and can help improve their performance. The experimental settings of the compared methods follow their original papers.

\textbf{Logit-based distillation}: Vanilla KD \cite{hinton2015distilling}, VBD \cite{hegde2020variational}, DTD-LA \cite{wen2021preparing}, DKD \cite{zhao2022decoupled}, CTKD \cite{li2023curriculum}.

\textbf{Feature-based distillation}: FitNet \cite{romero2014fitnets}, PKT \cite{passalis2018learning}, VID \cite{ahn2019variational}, SRRL \cite{yang2021knowledge} and SemCKD \cite{wang2022semckd}

\subsubsection{Training Details}
To ensure fair comparison, we follow the same training settings as in the previous works \cite{tiancontrastive,chen2021cross}. In all experiments, stochastic gradient descent (SGD) with momentum 0.9 is adopted as the parameter optimizer.
For CIFAR-100, the initial learning rate is set to 0.05 except for ShuffleNet and MobileNet, whose initial learning rate is 0.01 instead. The learning rate is divided by 10 at the 150, 180 and 210 epoch during the whole training process of 240 epochs. The batch size for both training and test sets is 64. 
For ImageNet, the model is trained 120 epochs, with an initial learning rate of 0.1, which is divided by 10 every 30 epochs. The batch size on ImageNet is set to 256.

Throughout the experiments, we set the temperature $\tau$ of vanilla KD loss to 4, which is consistent with \cite{tiancontrastive,chen2021cross}. In terms of DS, we select 80\% samples with higher influence scores as $D^t$ to be supervised by the teacher and the remaining 20\% samples as $D^s$ to be directly supervised by ground-truth labels. When conducting LR, the hyper-parameter $\eta$ is set as 0.8. We further explore the impact of hyper-parameters in the ablation study. The results on CIFAR-100 are reported as the mean of three trials, while the results on ImageNet are obtained from a single trial. 
As for hardware, the experiments on CIFAR-100 are conducted on a single NVIDIA Tesla P40, which three Tesla V100S are used for ImageNet.

\begin{table}[t]
\small
\centering
\caption{Comparison between influence-based selection and random selection on CIFAR-100. The teacher and student are ResNet32x4 and ResNet8x4, respectively.}
\centering
\renewcommand\arraystretch{1.35}
\resizebox{0.38\textwidth}{!}{
\setlength\tabcolsep{8pt} 
\begin{tabular}{p{0.9cm}<{\centering}|c|c}
\bottomrule[0.7pt]%\hline
\multicolumn{3}{c}{Baseline Acc: 74.12\%}\\
\hline
PCT(\%) & Strategy & Acc(\%)\\
\hline
\multirow{3}*{20} & Random & 70.54\\
& \begin{minipage}[b]{0.3\columnwidth}
    \centering
    \raisebox{-.2\height}
    {\includegraphics[width=1\textwidth]{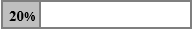}}
\end{minipage} & 70.10\\
& \begin{minipage}[b]{0.3\columnwidth}
    \centering
    \raisebox{-.2\height}
    {\includegraphics[width=1\textwidth]{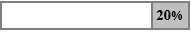}}
\end{minipage} & 71.13\\ 
\hline
\multirow{3}*{50} & Random & 73.05\\
& \begin{minipage}[b]{0.3\columnwidth}
    \centering
    \raisebox{-.2\height}
    {\includegraphics[width=1\textwidth]{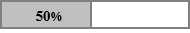}}
\end{minipage} & 73.60\\
&\begin{minipage}[b]{0.3\columnwidth}
    \centering
    \raisebox{-.2\height}
    {\includegraphics[width=1\textwidth]{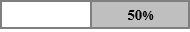}}
\end{minipage} & 73.69\\
\hline
\multirow{3}*{80} & Random & 74.70\\
& \begin{minipage}[b]{0.3\columnwidth}
    \centering
    \raisebox{-.2\height}
    {\includegraphics[width=1\textwidth]{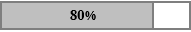}}
\end{minipage} & 74.59\\
&\begin{minipage}[b]{0.3\columnwidth}
    \centering
    \raisebox{-.2\height}
    {\includegraphics[width=1\textwidth]{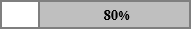}}
\end{minipage} & 74.81\\
% \bottomrule[0.7pt]
\hline
\end{tabular}}
\begin{tablenotes}
\footnotesize
\item{1.} The entire dataset is represented as a rectangle, where the samples are arranged in ascending order of influence score. 
\item{2.} The gray part represents $D^t$ and the white part is $D^s$. For example, the icon \begin{minipage}[b]{0.15\columnwidth} 
    % \centering
    {\includegraphics[width=1\textwidth]{Figure/DS1.png}}
\end{minipage} means that $D^t$ contains 20\% samples with lower score and $D^s$ contains the remaining 80\% samples with higher score.
\end{tablenotes}
\label{ds}
\end{table}

\begin{table}[t]
\small
\centering
\caption{Accuracy on CIFAR-100 with or without DS and LR. The teacher and student are ResNet32x4 and ResNet8x4, respectively. $\Delta$ refers to the performance improvement compared with baseline.}
\centering
\renewcommand\arraystretch{1.13}
\setlength\tabcolsep{10pt}
\begin{tabular}{c|cccc}
% \hline
\bottomrule[0.7pt]
Method & DS & LR & Acc (\%) & $\Delta$\\
\hline
\multirow{4}*{KD \cite{hinton2015distilling}} & $\times$  & $\times$ (KL) & 74.12 & -\\
& $\times$ & $\times$ (MSE) & 74.34 & 0.22\\
& $\times$ & \checkmark & 75.33 & 1.21\\
& \checkmark & $\times$ & 74.81 & 0.69\\
& \checkmark & \checkmark & \textbf{75.76} & \textbf{1.64}\\
\hline
\multirow{4}*{PKT \cite{passalis2018learning}} & $\times$  & $\times$  & 74.81 & -\\
& $\times$ & \checkmark & 75.15 & 0.34\\
& \checkmark & $\times$ & 74.93 & 0.12\\
& \checkmark & \checkmark & \textbf{75.53} & \textbf{0.72}\\

\bottomrule[0.7pt]
\end{tabular}
\label{simple}
\end{table}

\begin{table*}[!t]
% \small
\centering
\caption{Comparison with vanilla KD on the CIFAR-100. $\Delta$ is the performance improvement.}
\centering
\renewcommand\arraystretch{1.35}
\resizebox{\textwidth}{!}{
\setlength\tabcolsep{4.5pt}
\begin{tabular}{c|c|c|c|c|c|c|c|c|c|c}
% \hline
\bottomrule[0.7pt]
\multirow{2}*{Teacher} &  ResNet32$\times$4 & WRN-40-2& WRN-40-2 & ResNet56 & VGG13 &ResNet32$\times$4 & ResNet32$\times$4& ResNet32$\times$4 & WRN-40-2 & WRN-40-2\\
& 79.42 & 76.31 & 76.31 & 72.41 & 74.64 & 79.42 & 79.42 & 79.42 & 76.31 & 76.31\\
\multirow{2}*{Student} &  ResNet8$\times$4 &WRN-40-1 & WRN-16-2 & ResNet20 & VGG8 & VGG8 & ShuffleNetV1 &ShuffleNetV2 & MobileNetV2 & ShuffleNetV1\\
&73.09 & 71.92 & 73.51 & 69.06 & 70.36 & 73.09 & 71.92 & 73.51 & 69.06 & 70.36\\
\hline
Vanilla KD \cite{hinton2015distilling} & 74.12 & 73.42&74.92&70.66&72.66&72.73&74.07&74.45&69.07&74.83\\
Our & 76.60 & 74.53 & 75.99&71.61&74.31&74.08&75.26&76.78&69.39&76.90\\
$\Delta$ & \textbf{2.48} & \textbf{1.11} & \textbf{1.07} & \textbf{0.95} & \textbf{1.65} & \textbf{1.35} & \textbf{1.19} & \textbf{2.33} & \textbf{0.32} & \textbf{2.07}\\

\bottomrule[0.7pt]
\end{tabular}}
\label{comparison_kd}
\end{table*}

\begin{table*}[!t]
% \small
\centering
\caption{Comparison results on the CIFAR-100. Teachers and students have similar architectures. ``OA" refers to the original accuracy (\%), and ``LDA" represents the new accuracy (\%) after applying LR or DS, $\Delta$ is the performance improvement.}
\centering
\renewcommand\arraystretch{1.35}
\resizebox{\textwidth}{!}{
\setlength\tabcolsep{4.5pt}
\begin{tabular}{cc|ccc|ccc|ccc|ccc|ccc}
% \hline
\bottomrule[0.7pt]
\multirow{4}*{Type} & \multirow{2}*{Teacher} &  \multicolumn{3}{c|}{ResNet32$\times$4} & \multicolumn{3}{c|}{WRN-40-2}& \multicolumn{3}{c|}{WRN-40-2} & \multicolumn{3}{c|}{ResNet56} & \multicolumn{3}{c}{VGG13}\\
& & \multicolumn{3}{c|}{79.42} & \multicolumn{3}{c|}{76.31} & \multicolumn{3}{c|}{76.31} & \multicolumn{3}{c|}{72.41} & \multicolumn{3}{c}{74.64}\\
& \multirow{2}*{Student} &  \multicolumn{3}{c|}{ResNet8$\times$4} &\multicolumn{3}{c|}{WRN-40-1} & \multicolumn{3}{c|}{WRN-16-2} & \multicolumn{3}{c|}{ResNet20} & \multicolumn{3}{c}{VGG8}\\
& & \multicolumn{3}{c|}{73.09} & \multicolumn{3}{c|}{71.92} & \multicolumn{3}{c|}{73.51} & \multicolumn{3}{c|}{69.06} & \multicolumn{3}{c}{70.36}\\
\hline
\multicolumn{2}{c|}{-}& OA & LDA & $\Delta$ & OA & LDA & $\Delta$ & OA & LDA & $\Delta$ & OA & LDA & $\Delta$ & OA & LDA & $\Delta$\\
\hline
\multirow{4}*{Logits} & Vanilla KD \cite{hinton2015distilling} & 74.12 & 74.81 & 0.69  & 73.42 & 74.25 & 0.83 & 74.92 & 75.39 & 0.47& 70.66 & 71.32 & 0.66& 72.66 & 73.49 & 0.83\\
% & VBD \cite{hegde2020variational} & 74.31 & x & x & 73.62 & x & x & 75.10 & x & x& 71.13 & x & x& 73.21 & x & x\\
& DTD-LA \cite{wen2021preparing} & 73.78 & 75.15 & 1.37 & 73.49 & 73.76 & 0.27 & 74.73 & 75.54 & 0.81& 70.99 & 71.24 & 0.25 & 72.98 & 73.87 & 0.89\\
& DKD \cite{zhao2022decoupled} & 76.02 & 76.49 & 0.47 & 76.11 & 76.23 & 0.12 & 76.55 & 76.75 & 0.20 & 71.79 & 71.90 & 0.11 & 74.68 & 74.88 & 0.20\\
& CTKD \cite{li2023curriculum} & 74.49 & 75.24  & 0.75 & 73.84 & 74.21 & 0.37 & 75.51 & 75.72 & 0.21 & 71.13 & 71.99 & 0.86 & 73.36 & 73.84& 0.48\\
% & Our & 76.60 & - & - & 74.53 & - & - & 75.99 & - & -& 71.61 & - & -& 74.31 & - & -\\
\hline
\multirow{5}*{Features} & FitNet \cite{romero2014fitnets} & 74.32 &  75.72 & 1.40  & 74.12 & 74.56 & 0.44 & 75.04 & 75.68 & 0.64 & 71.52 & 71.96 & 0.44 & 73.54 & 73.86 & 0.32\\
& PKT \cite{passalis2018learning} & 74.81 & 75.53 & 0.72  & 73.51 & 73.78 & 0.27 & 75.60 & 75.76 & 0.16 & 70.92 & 71.35 & 0.43& 73.40 & 74.16 & 0.76\\
& VID \cite{ahn2019variational} & 74.49 & 75.90 & 1.41  & 74.20 & 74.79 & 0.59 & 74.79 & 75.14 & 0.35 & 71.71 & 72.01 & 0.30 & 73.96 & 73.61 & -0.35\\
& SRRL \cite{yang2021knowledge} & 75.39 & 76.15 & 0.76  & 74.98 & 75.16 & 0.18 & 75.55 & 76.20 & 0.65& 72.01 & 71.79 & -0.22& 74.68 & 74.81 & 0.13\\
& SemCKD \cite{wang2022semckd} & 75.58 & 76.35 & 0.77  & 74.78 & 74.57 & -0.21 & 75.42 & 75.52 & 0.10& 71.98 & 72.31 & 0.33 & 74.42 & 74.75 & 0.33\\
\bottomrule[0.7pt]
\end{tabular}}
\label{same}
\end{table*}

\begin{table*}[!t]
% \small
\centering                                                                                                      
\caption{Comparison results on the CIFAR-100. Teachers and students have different architectures. ``OA" refers to the original accuracy (\%), and ``LDA" represents the new accuracy (\%) after applying LR or DS, $\Delta$ is the performance improvement.}
\centering
\renewcommand\arraystretch{1.35}
\setlength\tabcolsep{4.5pt}
\resizebox{\textwidth}{!}{
\begin{tabular}{cc|ccc|ccc|ccc|ccc|ccc}
% \hline
\bottomrule[0.7pt]
\multirow{4}*{Type} & \multirow{2}*{Teacher} &  \multicolumn{3}{c|}{ResNet32$\times$4} & \multicolumn{3}{c|}{ResNet32$\times$4}& \multicolumn{3}{c|}{ResNet32$\times$4} & \multicolumn{3}{c|}{WRN-40-2} & \multicolumn{3}{c}{WRN-40-2}\\
& & \multicolumn{3}{c|}{79.42} & \multicolumn{3}{c|}{79.42} & \multicolumn{3}{c|}{79.42} & \multicolumn{3}{c|}{76.31} & \multicolumn{3}{c}{76.31}\\
% \cline{2-17}
& \multirow{2}*{Student} &  \multicolumn{3}{c|}{VGG8} &\multicolumn{3}{c|}{ShuffleNetV1} & \multicolumn{3}{c|}{ShuffleNetV2} & \multicolumn{3}{c|}{MobileNetV2} & \multicolumn{3}{c}{ShuffleNetV1}\\
& & \multicolumn{3}{c|}{73.09} & \multicolumn{3}{c|}{71.92} & \multicolumn{3}{c|}{73.51} & \multicolumn{3}{c|}{69.06} & \multicolumn{3}{c}{70.36}\\
\hline
\multicolumn{2}{c|}{-}& OA & LDA & $\Delta$ & OA & LDA & $\Delta$ & OA & LDA & $\Delta$ & OA & LDA & $\Delta$ & OA & LDA & $\Delta$\\
\hline
\multirow{4}*{Logits} & Vanilla KD \cite{hinton2015distilling} & 72.73 & 72.92 & 0.19  & 74.07 & 74.21 & 0.14 & 74.45 & 75.45 &  1.00 & 69.07 & 69.54 & 0.47 & 74.83 & 75.50 & 0.67\\
% & VBD \cite{hegde2020variational} & - & - & -  & 74.21 & x & x & - & - & -& 69.22 & x & x& 75.10 & x & x\\
& DTD-LA \cite{wen2021preparing} & 72.67 & 73.00 & 0.33  & 73.99 & 74.88 & 0.89 & 75.05 & 76.24 & 1.19 & 68.99 & 69.57 & 0.58& 74.90 & 75.87 & 0.97\\
& DKD \cite{zhao2022decoupled} & 74.10 & 74.55 & 0.45  & 75.88 & 75.71 & -0.17 & 76.87 & 77.06 & 0.19 & 69.47 & 69.58 & 0.11 & 76.41 & 76.52 & 0.11\\
& CTKD \cite{li2023curriculum} & 73.54 & 74.27 & 0.73 & 74.37 & 75.49& 1.12 & 75.42 & 75.51 & 0.09 & 69.21 & 69.45 & 0.24  & 75.80 & 76.14 & 0.34 \\
% & Our & 74.08 & - & - & 75.26 & - & - & 76.78 & - & - & 69.39 & - & -& 76.90 & - & -\\
\hline
\multirow{5}*{Features} & FitNet \cite{romero2014fitnets} & 72.91 & 73.53 & 0.62  & 74.52 & 74.64 & 0.12 & 74.23 & 75.42 & 1.19& 68.71 & 68.77 & 0.06 & 74.11 & 76.11 & 2.00\\
& PKT \cite{passalis2018learning} & 73.08 & 73.82 & 0.74  & 74.05 & 74.94 & 0.89 & 74.69 & 75.84& 1.15 & 68.80 & 69.06& 0.26 & 75.68 & 75.87 & 0.19\\
& VID \cite{ahn2019variational} & 73.19 & 74.04  & 0.85 & 74.28 & 75.58 & 1.30 & 75.22 & 76.01 & 0.79 & 68.91 & 68.33& -0.58 & 74.41 & 75.88 & 1.47\\
& SRRL \cite{yang2021knowledge} & 74.06 & 74.57 & 0.51  & 75.38 & 76.04 & 0.66 & 76.19 & 77.07 & 0.88 & 69.34 & 69.56 & 0.22 & 75.22 & 76.23 & 1.01\\
& SemCKD \cite{wang2022semckd} & 75.27 & 75.51 & 0.24 & 75.41 & 76.45 & 1.04 & 77.63 & 77.85 & 0.22 & 69.88 & 69.98 & 0.10 & 76.83 & 77.46 & 0.63\\
\bottomrule[0.7pt]
\end{tabular}}
\label{different}
\end{table*}

\subsection{Effect of LR and DS}
\textbf{Influenced-based selection v.s. Random selection.}
When conducting DS, we propose to select appropriate samples according to certain criterion (e.g., influence score), where the entire dataset $D$ is split into two parts $D^t$ and $D^s$. $D^t$ is input to both the teacher and student models at the same time, while $D^s$ is only applied to the student.
To investigate the effect of this kind of specific selection and random selection, which is the simplest way to realize DS, we evaluate the performance on CIFAR-100. For detailed illustration, we also set different selection percentages of $D^t$ (i.e., 20\%, 50\% and 80\%), and the baseline is vanilla KD. 

Table \ref{ds} presents the results of our experiments on the impact of the amount of data input to the teacher on the distillation accuracy. The results indicate that, when the amount of data input to the teacher is relatively small, the distillation accuracy is worse than vanilla KD. This finding highlights the necessity of the guidance from the teacher model. For instance, when only 20\% samples are used for the teacher, the accuracy is only around 71\%, which is significantly lower than vanilla KD. When there are enough samples to receive supervision from the teacher, the performance of selecting data based on influence score is relatively better than random selection. Furthermore, the accuracy gain of selecting data with higher scores is more significant. We ascribe this kind of phenomenon to an hypothesis that the samples with higher scores may be more difficult to be classified, so that they need supervision from teacher model to provide more information for classification. Thus, based on these observations, we follow the strategy of selecting 80\% samples with higher influence score to be supervised by teacher in the following experiments. 

\textbf{Performance gain of each part.} We also conduct a simple comparison on CIFAR-100 to evaluate the contribution brought by the proposed LR and DS. The results are presented in Table \ref{simple}. To explore the effect of LR and DS, we individually apply them on vanilla KD \cite{hinton2015distilling} and PKT \cite{passalis2018learning}. The results show that both DS and LR are helpful to improve the distillation performance, but the gain from individually applying one of them is limited (e.g., 0.12\% of DS and 0.34\% of LR on PKT). Therefore, the results suggest that the combination of DS and LR is necessary to achieve better performance. Besides, to eliminate the influence of MSE, we also compare the performance of directly replacing KL divergence with MSE for calculating the KD loss. It can be noted that the performance gain is still lower than using LR and DS, showing the effectiveness of the proposed method. Furthermore, it is initially illustrated that DS and LR are compatible with other distillation methods to improve their performance. More combination results will be provided in the following section.

\subsection{Main Results}
\textbf{CIFAR-100.} In Table \ref{comparison_kd}-\ref{different}, we provide a comprehensive comparison of our method with other distillation approaches, including both logits-based and feature-based methods. We also choose various combinations of the teacher and student models, where the architectures of models are either similar (Table \ref{same}) or quite different (Table \ref{different}).

\begin{figure}[!t]
\centering
\includegraphics[width=0.9\columnwidth]{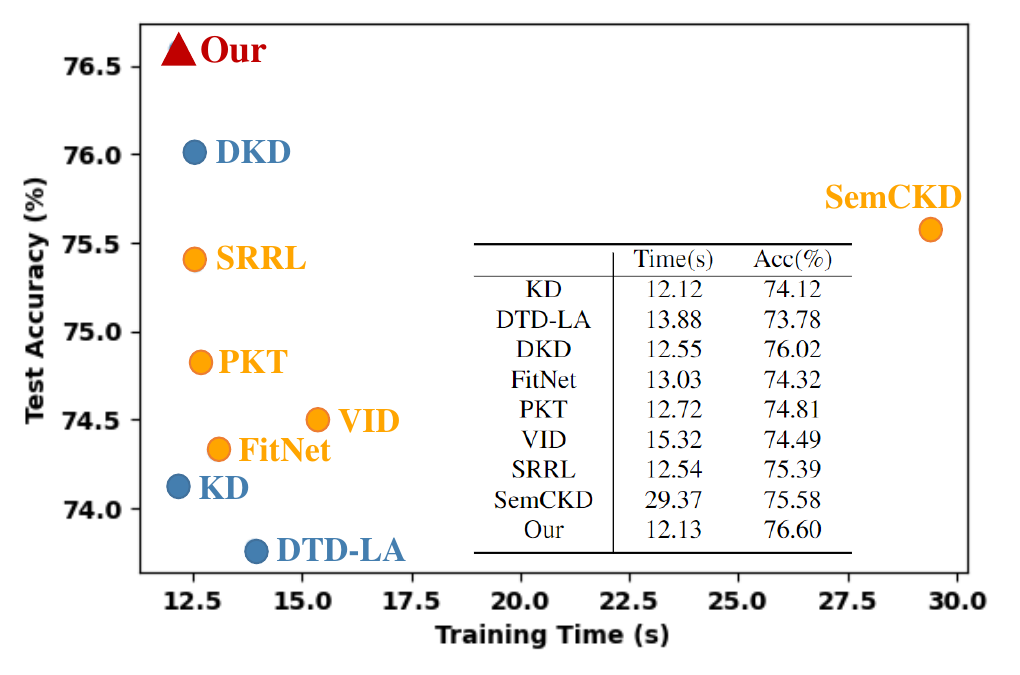}
\caption{Comparison of training time and accuracy on CIFAR-100.
The teacher and student are ResNet32x4 and ResNet8x4, respectively.}
\label{time}
\end{figure}

In Table \ref{comparison_kd}, the accuracy of our proposed method can be improved on all combinations of teacher and student compared with vanilla KD, and the improvement are significant in some cases. It is also inspired that our method can achieve comparable or even better performance than some of feature-based distillation. For example, for ``WRN-40-2" and ``WRN-16-2” pair, our method achieves 75.99\% accuracy, which is higher than other feature-based approaches. For the pair of ``ResNet32x4\_ResNet8x4" and ``WRN-40-2\_ShuffleNetV1", our method outperforms all competitors. Moreover, our method incurs a lower computation cost than feature-based methods, as the latter often requires tedious computation to transform intermediate feature maps that may hinder their application on resource-constrained devices. We assess the training time of different methods on CIFAR-100 in Figure \ref{time}, and the results demonstrate that our method achieves a better trade-off between the training efficiency and accuracy than feature-based approaches.

To further evaluate the effectiveness of the proposed LR and DS, we also apply them to other distillation approaches, where the hyper-parameters are set as the same as original papers. For logits-based methods, only DS technique is combined because the logits loss has already been modified. For feature-based methods, both LR and DS are applied. The results show that in most cases, the distillation performance can be further boosted through applying DS and LR. For instance, the original accuracy of VID \cite{ahn2019variational} on the pair of ``WRN-40-2" and ``ShuffleNetV1" is only 74.41\%, which is even lower than 74.83\% of vanilla KD. However, the accuracy is increased by 1.47\% after applying DS and LR, outperforming the vanilla KD by 1.05\%. These results strongly demonstrate the effectiveness of our proposed LR and DS, and also illustrate an encouraging property that LR and DS are highly compatible with the state-of-the-art distillation methods.

\begin{table}[t]
\small
\centering     
\caption{Comparison results on ImageNet. The accuracy metrics are Top-1 values (\%).}
\centering
\renewcommand\arraystretch{1.35}
\setlength\tabcolsep{10pt}
\begin{tabular}{c|c|c}
% \hline
\bottomrule[0.7pt]
\multirow{2}*{Teacher} &  ResNet34 & ResNet50\\
& 73.31 & 76.26\\
% \cline{2-17}
\multirow{2}*{Student} &  ResNet18 & ResNet18\\
& 70.04 & 70.04\\
\hline
Vanilla KD \cite{hinton2015distilling}& 70.66 & 71.29\\
LR & 70.83 & 71.36\\
CTKD \cite{li2023curriculum} & 71.22 & 71.31 \\
VID \cite{ahn2019variational} & 70.30& 71.11\\
SRRL \cite{yang2021knowledge}&70.95  &71.46 \\
SemCKD \cite{wang2022semckd}&70.87 & 71.41\\
\hline

SRRL+LR & 71.10  & 71.58 \\
CTKD+LR & 71.30 & 71.43 \\
\bottomrule[0.7pt]
\end{tabular}
\label{imagenet}
\end{table}

\textbf{ImageNet.} We evaluate the performance of our proposed LR technique on ImageNet using two popular teacher-student model pairs, i.e., ``ResNet34-ResNet18" and ``ResNet50-ResNet18". The results are presented in Table \ref{imagenet}. Compared to vanilla KD, our LR achieves encouraging improvement on Top-1 accuracy, further narrowing the gap between the student and teacher models. This also validates that our method is effective on large-scale datasets. Additionally, we apply LR on other approachs such as SRRL and CTKD, and it obtains more favourable performance compared to other competitors, verifying the compatibility of our proposed method again.

\begin{table}[t]
\small
\centering
\caption{Analysis of coefficient $\lambda_1$ and $\lambda_2$ on CIFAR-100. When changing one of them, We fix the other one as 1 for simplification.}
\centering
\renewcommand\arraystretch{1.25}
\setlength\tabcolsep{2.5pt}
\begin{tabular}{c|ccccccccc}
% \hline
\bottomrule[0.7pt]
\multicolumn{10}{c}{Baseline: 74.81\%}\\
\hline
$\lambda_1$ & 0 & 0.1 & 0.5 & 0.8 &1 & 2 & 4 & 8 & 10\\
% 0.3(74.86)
% \hline
Acc(\%) & 72.96 & 73.73 & 75.02 & 75.30 & 75.76 & 75.75 &\textbf{76.60} & 76.54 & 76.46\\
\hline
$\lambda_2$ & 0&0.1 & 0.5 & 0.8 &1 & 2 & 4 & 8 & 10\\
% \hline
% 0.3 (75.48)
Acc(\%) & 75.01 &\textbf{76.31} & 75.09 & 75.34 & 75.76 & 75.54 & 75.20 & 75.07 & 75.45\\
\bottomrule[0.7pt]
\end{tabular}
\label{lambda}
\end{table}

\begin{table}[!t]
\small
\centering
\caption{Different setting of $\eta$ on CIFAR-100. We select different values of $\eta$ including fixed and learnable values.}
\centering
\renewcommand\arraystretch{1.25}
\setlength\tabcolsep{4pt}
\begin{tabular}{c|cccccc|cc}
% \hline
\bottomrule[0.7pt]
\multicolumn{9}{c}{Baseline: 74.81\%}\\
\hline
\multirow{2}*{$\eta$} & \multicolumn{6}{c|}{Fixed} & \multicolumn{2}{c}{Learnable}\\
\cline{2-9}
& 0.5 & 0.7 & 0.8 & 0.85 & 0.9 & 0.95 & $p^t_{max}$ & $p^t_{tar}$ \\
\hline
Acc(\%) & 75.45 & 75.58 & 75.76 & 75.59 & 75.18 & 75.03 & 75.34 & 75.6\\

\bottomrule[0.7pt]
\end{tabular}
\begin{tablenotes}
\footnotesize
\item{1.} For learnable values, we set $p^t_{max}$ and $p^t_{tar}$, that is, the probability of predicted class and target class in teacher's prediction, respectively.
\end{tablenotes}
\label{eta}
\end{table}

\subsection{Additional Analysis}

\textbf{Analysis of Coefficients $\lambda_1$ and $\lambda_2$.} 
We also evaluate the impact of coefficients $\lambda_1$ and $\lambda_2$ in Eq. (\ref{eq:wrong_right}), which are introduced to balance each loss term. For evaluation, we set various values of $\lambda_1$ and $\lambda_2$ ranging from 0 to 10, that is $\lambda_1, \lambda_2 \in \{0.1, 0.3, 0.5, 0.8, 1, 2, 4, 8, 10\}$. The results are showed in Table \ref{lambda}. Our method outperforms baseline in most of the cases, showing the effectiveness again. It is also indicated that both the logits loss of the right and wrong parts are indispensable, the absence of either will result in a sharp decline of accuracy, especially when $\lambda_1 = 0$. As $\lambda_1$ rises from 0 to 10, the accuracy first increases gradually and reaches a maximum at the point of ${\lambda_1} = 4$. 
As for $\lambda_2$, the performance gain are relatively stable with different $\lambda_2$ around 1.

\textbf{Analysis of Coefficient $\eta$.} We explore the sensitivity of the hyper-parameter $\eta$ in Eq. (\ref{eq:eta}). Table \ref{eta} reports the performance with different $\eta$ on CIFAR-100. Since $\eta$ is a coefficient between 0 and 1, we first choose some fixed values as \{0.5, 0.7, 0.8, 0.85, 0.9, 0.95\}. In addition to these fixed values, we also set $\eta$ as a learnable parameter that varies for different input samples. For example, we directly regard the probabilities of predicted class ($p^t_{max}$) and target class ($p^t_{tar}$) in teacher's prediction as $\eta$, whose value ranges are also 0-1. Here, other hyper-parameters $\lambda_1$ and $\lambda_2$ are set as 1. It can be observed that our method shows its superiority to baseline under varying $\eta$, where the performance gain ranging from 0.22\% to 0.95\%, and it achieves the best performance when $\eta$ is set to around 0.8. From another perspective, the fluctuation of accuracy is relatively small, demonstrating the robustness of our method to hyper-parameter $\eta$.

\section{Conclusion}
\textbf{Conclusions.}
Knowledge distillation has been hampered by the issue of incorrect supervision from the teacher model. In this paper, we have proposed to alleviate the impact of such incorrect supervision from two aspects, which are simple but effective. Firstly, we have proposed LR to rectify the wrong predictions of the teacher according to the ground truth. Secondly, we have also introduced DS to select appropriate samples to be supervised by the teacher. Experiments on both small and large-scale datasets have been conducted to justify the effectiveness of the proposed LR and DS. The statistical results have demonstrated that our proposed method achieves better performance than vanilla KD or even feature-based methods, and is more efficient for training without tedious computation to transfer features.
Furthermore, as a plug-in technique, our method can be easily combined with other distillation approaches that can further improve their performance.

\textbf{Limitations and Future Work.} 
For LR, we rectify incorrect predictions by combining the correct information contained in the ground truth, with an assumption that the training samples are labeled correctly. However, in real-world applications, ground truth labels are sometimes incomplete or missing, which limits the effectiveness of LR. Therefore, revising wrong supervision without relying on ground truth deserves further exploration.

Regarding DS, this paper only uses the influence function to estimate the values of each sample. Other estimation methods are also worth investigating in the future, as they may help select more appropriate samples for distillation to further enhance performance. Additionally, the current method of estimating values needs to work on each sample, which would be time-consuming for large-scale datasets. Designing more efficient approaches is also a direction of future work.

\bibliographystyle{IEEEtran}
\bibliography{ref}

\end{document}